\documentclass[runningheads]{llncs}

\usepackage{eccv}

\usepackage{eccvabbrv}

\usepackage{graphicx}
\usepackage{booktabs}

\usepackage[accsupp]{axessibility}  

\usepackage[pagebackref,breaklinks,colorlinks,citecolor=eccvblue]{hyperref}

\usepackage{orcidlink}
\usepackage{algorithm}
\usepackage{algpseudocode}
\usepackage{multicol}
\usepackage{multirow}
\usepackage{makecell}
\usepackage{subcaption}
\usepackage{chngcntr}
\usepackage[symbol]{footmisc}
\usepackage[dvipsnames]{xcolor}

\newcommand*{\mline}[1]{%
\begingroup
    \renewcommand*{\arraystretch}{1.1}%
   \begin{tabular}[c]{@{}>{\raggedright\arraybackslash}p{2cm}@{}}#1\end{tabular}%
  \endgroup
}
\begin{document}

\newcommand{\ourwork}{InceptionHuman}
\newcommand{\bG}[0]{{\bf G}}
\newcommand{\bd}[0]{{\bf d}}
\newcommand\tab[1][1cm]{\hspace*{#1}}

\title{InceptionHuman: Controllable Prompt-to-NeRF for Photorealistic 3D Human Generation}
\titlerunning{InceptionHuman}

\author{Shiu-hong Kao\inst{1} \and
Xinhang Liu\inst{1} \and
Yu-Wing Tai\inst{2} \and Chi-Keung Tang\inst{1}}

\authorrunning{S. Kao et al.}

\institute{The Hong Kong University of Science and Technology\and
Dartmouth College\\
\email{\{skao,xliufe\}@connect.ust.hk, yuwing@gmail.com, cktang@cs.ust.hk}}

\maketitle

\begin{abstract}
This paper presents {\sc \ourwork}\footnote{{\em Inception} (2010) demonstrates the impact of ``shared dreaming''
in the movie.
Likewise, the ensemble of human images hallucinated by {\sc \ourwork} works together in the shared NeRF space to produce a 3D-consistent human.
}, a prompt-to-NeRF framework that allows easy control via a combination of prompts in different modalities (e.g., text, poses, edge, segmentation map, etc) as inputs to generate photorealistic 3D humans. 
While many works have focused on generating 3D human models, they suffer one or more of the following: lack of distinctive features, unnatural shading/shadows, unnatural poses/clothes, limited views, etc.
\ourwork~achieves consistent 3D human generation within a progressively refined NeRF space with two novel modules, \emph{Iterative Pose-Aware Refinement} (IPAR) and \emph{Progressive-Augmented Reconstruction} (PAR). IPAR iteratively refines the diffusion-generated images and synthesizes high-quality 3D-aware views considering the close-pose RGB values. PAR employs a pretrained diffusion prior to augment the generated synthetic views and adds regularization for view-independent appearance.
Overall, the synthesis of photorealistic novel views empower the resulting 3D human NeRF from 360 degree perspectives. Extensive qualitative and quantitative experimental comparison show that our InceptionHuman models achieve state-of-the-art application quality. 

  \keywords{Neural radiance field \and Diffusion model \and 3D human}
\end{abstract}

\section{Introduction}
\label{sec:intro}
While generative models have achieved significant success in multiple domains, such as motion~\cite{tevet2022human}, mesh~\cite{michel2022text2mesh,liu2023meshdiffusion}, and 2D images~\cite{gan,vae,vqvae,stablediffusion,ddpm}, creation of realistic 3D contents remains challenging, especially for articulated objects like clothed human bodies. 

\begin{figure}[t]
    \centering
    \captionsetup[subfigure]{labelformat=empty}
    \resizebox{\textwidth}{!}{%
    \begin{subfigure}{0.16\linewidth}
        \centering
        \includegraphics[width=\textwidth]{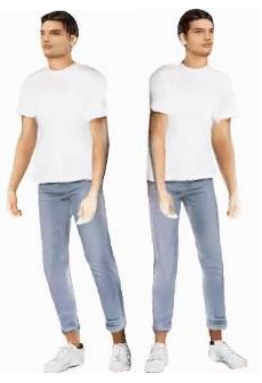}
        \caption{EVA3D~\cite{eva3d}}
        \label{fig:intro_eva3d}
    \end{subfigure}
    \hfill
    \begin{subfigure}{0.2\linewidth}
        \centering
        \includegraphics[width=.8\textwidth]{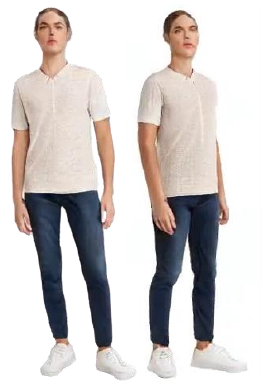}
        \caption{3DHumanGAN~\cite{yang20233dhumangan}}
        \label{fig:intro_3dhumangan}
    \end{subfigure}
    \hfill
    \begin{subfigure}{0.21\linewidth}
        \centering
        \includegraphics[width=\textwidth]{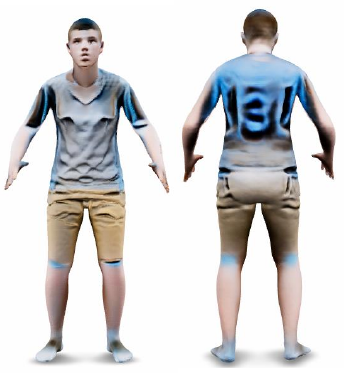}
        \caption{AvatarCLIP~\cite{avatarclip}}
        \label{fig:intro_avatarclip}
    \end{subfigure}
    \hfill
    \begin{subfigure}{0.21\linewidth}
        \centering
        \includegraphics[width=.78\textwidth]{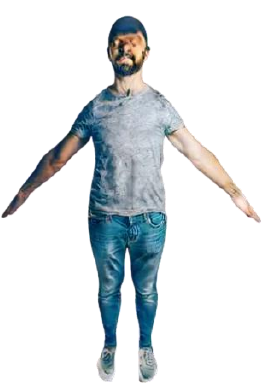}
        \caption{DreamAvatar~\cite{cao2023dreamavatar}}
        \label{fig:intro_dreamavatar}
    \end{subfigure}
    \hfill
    \begin{subfigure}{0.13\linewidth}
        \centering
        \includegraphics[width=.83\textwidth]{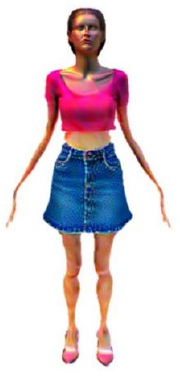}
        \caption{TADA~\cite{tada}}
        \label{fig:intro_tada}
    \end{subfigure}
    \hfill
    \begin{subfigure}{0.29\linewidth}
        \centering
        \includegraphics[width=\textwidth]{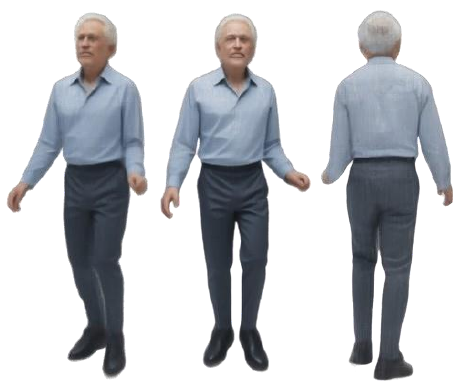}
        \caption{InceptionHuman (Ours)}
        \label{fig:intro_ours}
    \end{subfigure}
    }
    \caption{Prior works on realistic 3D human generation generally suffer from lack of distinctive features~\cite{eva3d}, unnatural clothes and shadows~\cite{avatarclip,cao2023dreamavatar,tada}. Some of them can only synthesize images from limited views~\cite{eva3d,yang20233dhumangan}. \ourwork~is a prompt-driven model for photorealistic 3D human generation that overcomes these limitations.}
    \label{fig:sota}
\end{figure}

Inspired by recent breakthrough in 2D image generative models, one straightforward and plausible approach  for 3D object generation is to train such models, \eg, GAN~\cite{gan} or DDPM~\cite{ddpm}, with 3D voxels or triplanes. However, this approach is currently impractical due to the limited diversity of 3D datasets. To address this issue, existing studies have attempted various methods by decomposing 3D generations into low-level subproblems. For example, \cite{avatarclip,humannorm,eva3d,pifu,yang20233dhumangan} decouple the problem into geometry generation and texture generation.
Although these approaches, particularly exemplified in the recent advances in HumanNorm~\cite{humannorm}, have demonstrated impressive results, they hinge on a simplified graphic rendering model, synthesizing inconsistent images with unnatural shapes, unnatural shadows, and lack of distinctive features, as shown in Figure~\ref{fig:sota}. Consequently, the rendered appearances lack photorealism in many ways.
One robust way to construct a 3D object with 3D scene representation is utilizing Neural Radiance Fields~\cite{nerf}, which models the 3D rendering problem with Multi-layer Perceptrons (MLP). NeRF is commonly used as a tool of novel view synthesis in which the goal is to render arbitrary unseen viewpoints of a scene with a given set of captured images and their associated camera poses. Prior works \cite{dreamhuman,sherf,zeroavatar} have striven to incorporate the generative models, \eg, diffusion models~\cite{stablediffusion}, with NeRF, but the results are still far from perfect. This is because vanilla NeRF requires a large number of training views and is prone to generating severe artifacts when the training views are sparse, blurry, inconsistent, or low-quality. In particular, previous works such as \cite{dreamhuman}, which incorporate diffusion models, generally suffer from the inconsistency issue as a consequence of the decoupled synthetic framework. 

\begin{figure}[t]
    \centering
\includegraphics[width=\linewidth]{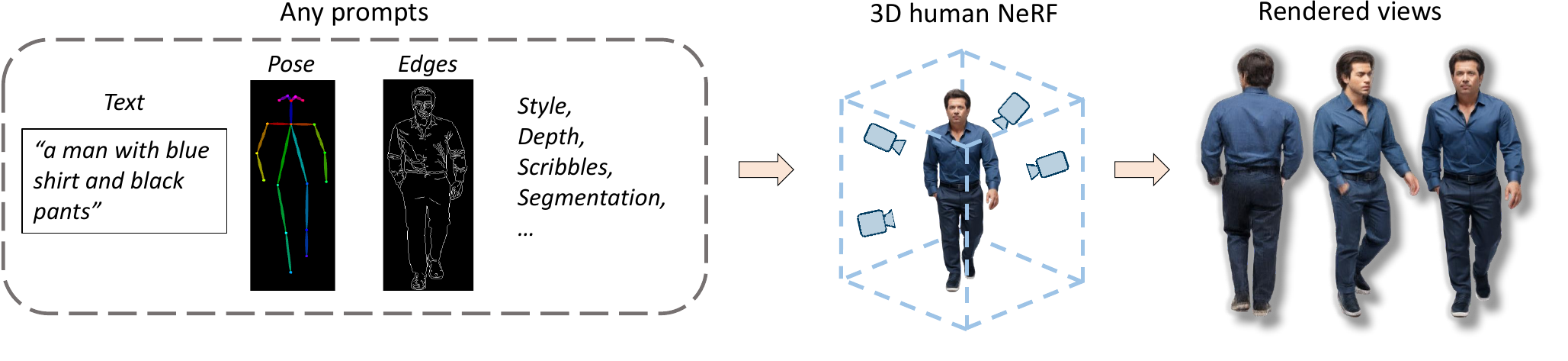}
    \caption{\ourwork~can receive different types of prompts as input and generates a high-quality NeRF-based 3D human. “Any prompts” refer to prompts in many varieties as listed.}
    \label{fig:impression}
    \vspace{-1.2em}
\end{figure}

{\sc \ourwork}~is a 3D human generative framework that combines conditional 2D diffusion models~\cite{stablediffusion} and NeRF to achieve prompt-guided generation, as shown in Figure~\ref{fig:impression}. \ourwork~leverages the advantages of conditional diffusion models that accept different types of controls, e.g., text, style, pose, edges, depth, seed images, etc., by using Latent Diffusion Model~\cite{stablediffusion} and ControlNet~\cite{controlnet}. 
To ensure consistency,  synthetic 2D human images must work collaboratively in the shared NeRF space to generate a consistent 3D human, where the corresponding 2D projections should agree with the synthetic images.  To this end, in~\ourwork, we employ two modules, \emph{Iterative Pose-Aware Refinement} (IPAR) and \emph{Progressive-Augmented Reconstruction} (PAR), to connect the 2D generative models and 3D reconstruction models for progressive refinement.

In particular, 
to generate consistent 2D views for NeRF reconstruction, InceptionHuman first uses IPAR to iteratively refine the diffusion-generated images in a frame-by-frame manner, where views with close poses are assumed to have similar RGB values. PAR is then employed, which leverages a pretrained diffusion prior, to progressively augment the consistent images and adding view-independent regularization. With the above consistency control, 
our method can discard the view-dependent components during the generation of high-quality images, which are undesirable high-frequency colors jittering along the views. Extensive experiments show that our InceptionHuman can generate high-quality 3D human bodies with arbitrary controls, and outperforms the state-of-the-art qualitatively and quantitatively.

To summarize, our contributions are as follows:
\begin{itemize}
    \item We propose a novel pipeline to connect diffusion models and NeRF and to generate realistic 3D human.
    \item We introduce IPAR and PAR to enhance the consistency among diffusion-synthesized images.
    \item We demonstrate the first 3D human model that accepts inputs with any controls, e.g., text, pose, style, edges, depth, seed images, etc.
\end{itemize}

\section{Related work}
\label{sec:related}
Our work is related to image generation~\cite{autoregression,vae,vqvae,gan,ddpm,improved_ddpm}, novel view synthesis~\cite{nerf,mipnerf,nerfies,jiang2020sdfdiff,kellnhofer2021neural,liu2020neural,martin2021nerf,park2019deepsdf} and 3D generative tasks~\cite{dreamfusion,hollein2023text2room,devries2021unconstrained,schwarz2020graf}. In this paper, we focus on high-quality text-to-3D human generation using the NeRF representation.

\vspace{2mm}
\noindent{\textbf{Novel view synthesis with NeRF.}} Neural Radiance Fields (NeRFs)~\cite{nerf} have become the mainstream technique for novel view synthesis, which leverages continuous 3D fields and volumetric rendering to synthesize novel views given a set of 2D images. Follow-up works further enhance the rendering quality and expand its applications under different scenarios, such as dynamic scene reconstruction~\cite{zhang2021editable,nerfies,dnerf,tretschk2021non}, 3D segmentation~\cite{rcnn, zarzar2022segnerf, liu2022unsupervised}, object detection~\cite{nerfrpn,nerfdet}, sparse view reconstruction~\cite{regnerf,deceptivenerf,dietnerf,freenerf,roessle2022dense}, 3D scene editing~\cite{fdnerf,liu2021editing,zhang2021editable,clipnerf}, and acceleration~\cite{yu2021plenoctrees, tensorf, instantngp, wang2022fourier}. Our \ourwork~can progressively enhance a NeRF model, cooperating with state-of-the-art generative 2D models to hallucinate individual but relevant and highly realistic human images, for inferring high-quality, 3D-consistent views of a human.

\vspace{2mm}
\noindent{\textbf{2D Diffusion models.}} Denoising Diffusion Probabilistic Models (DDPMs~\cite{ddpm}) is a powerful class of generative models that follows a Markov process to generate high-quality 2D images by denoising input images. Recent advancement has shown that DDPMs achieve notable success in conditional generation, \emph{e.g.,} text-to-image generation~\cite{stablediffusion,podell2023sdxl}, inpainting~\cite{lugmayr2022repaint,saharia2022palette}, and spatial image controls~\cite{controlnet}. Our work presents a 3D generation framework, adopting  a conditional DDPM for 3D-aware pseudo-view generation and denoising with view-consistent loss.

\vspace{2mm}
\noindent{\textbf{3D human/avatar generation.}} 
The recent improvements in human pose estimation~\cite{mcnally2021evopose2d,raaj2019efficient,cao2017realtime} and mesh reconstruction~\cite{econ,joo2018total,xiu2022icon} have enabled the creation of detailed 3D realistic human~\cite{sherf,humannorm,eva3d,pifu} and cartoon-styled avatar~\cite{magicavatar,zeroavatar,avatarclip,dreamwaltz,dreamhuman}. In~\cite{sherf,zeroavatar,pifu}, 3D human/avatar can be constructed from a single image by extracting the representations in canonical space. EVA3D~\cite{eva3d} introduces a GAN-based framework and divides NeRFs into local parts to generate unconditional 3D human. DreamGaussian~\cite{dreamgaussian} enables general 3D subject generation by utilizing a 3D Gaussian Splatting model that converts 3D Gaussians into extracts textured meshes and applies a fine-tuning stage to enhance the details. Closer to our work, some recent studies have attempted to generate 3D human from a text-guided aspect. DreamWaltz~\cite{dreamwaltz} extracts pose priors from a SMPL~\cite{loper2023smpl} mesh input to generate cartoon-styled avatar. AvatarCLIP~\cite{avatarclip} incorporates SMPL and Neus~\cite{wang2021neus} for 3D human representations, leveraging CLIP as supervision to achieve zero-shot text-guided avatar generation. These methods, while successfully generate 3D human/avatar bodies, still have much room for improvement.

Recent studies also approach the human/avatar generation problem with the assistance of diffusion models~\cite{ddnerf,jiang2022neuman,instructnerf2023}. 
For text-guided 3D human generation, one close work is DreamHuman~\cite{dreamhuman} that incorporates diffusion models with a pose-conditioned NeRF constrained on ImGHUM~\cite{alldieck2021imghum} to generate animatable NeRF-based 3D characters. HumanNorm~\cite{humannorm} decouples the generation problem into two stages, by first training 3D depth-adapted and normal-adapted diffusion models for geometry, and then generating texture by introducing a normal-aligned diffusion model. These methods show a high-quality 3D human at first sight, yet suffer from the simplified graphic rendering model, resulting in appearances that lack realism, due to the decoupling process. In this paper, we aims to connect the 2D diffusion models with NeRF to generate high-quality 3D-aware human bodies.

\section{InceptionHuman Framework}\label{sec:method}

\begin{figure}[t]
    \centering
\begin{subfigure}{0.457\linewidth}
    \includegraphics[width=\textwidth]{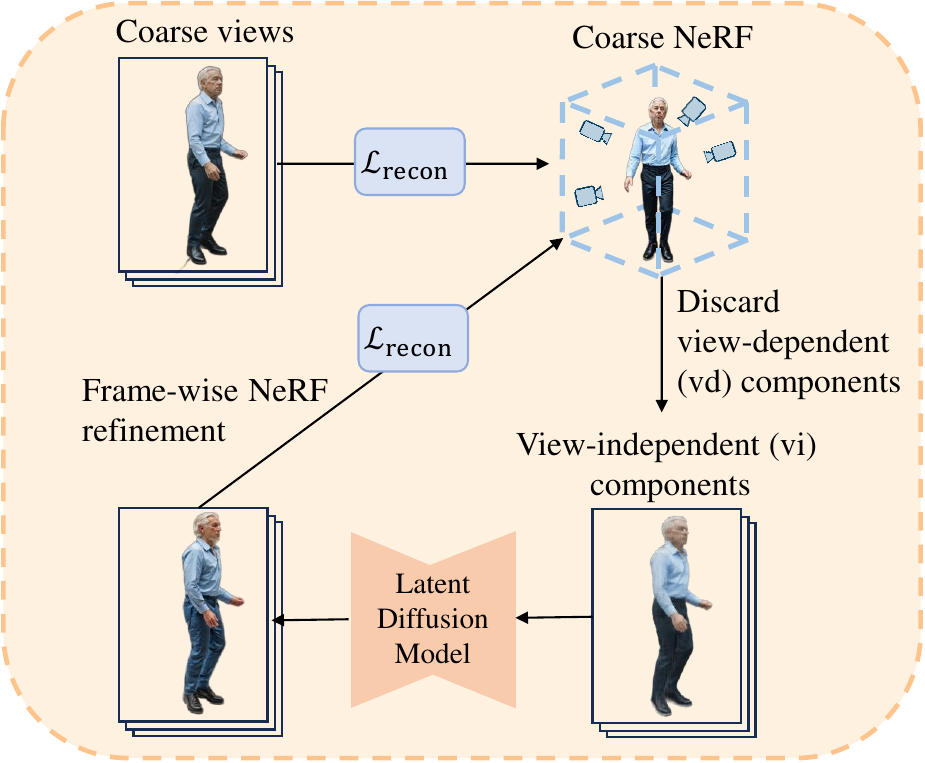}
    \caption{Iterative Pose-Aware Refinement.}
    \label{fig:ipar}
\end{subfigure}
\hfill
\begin{subfigure}{0.525\linewidth}
    \includegraphics[width=\textwidth]{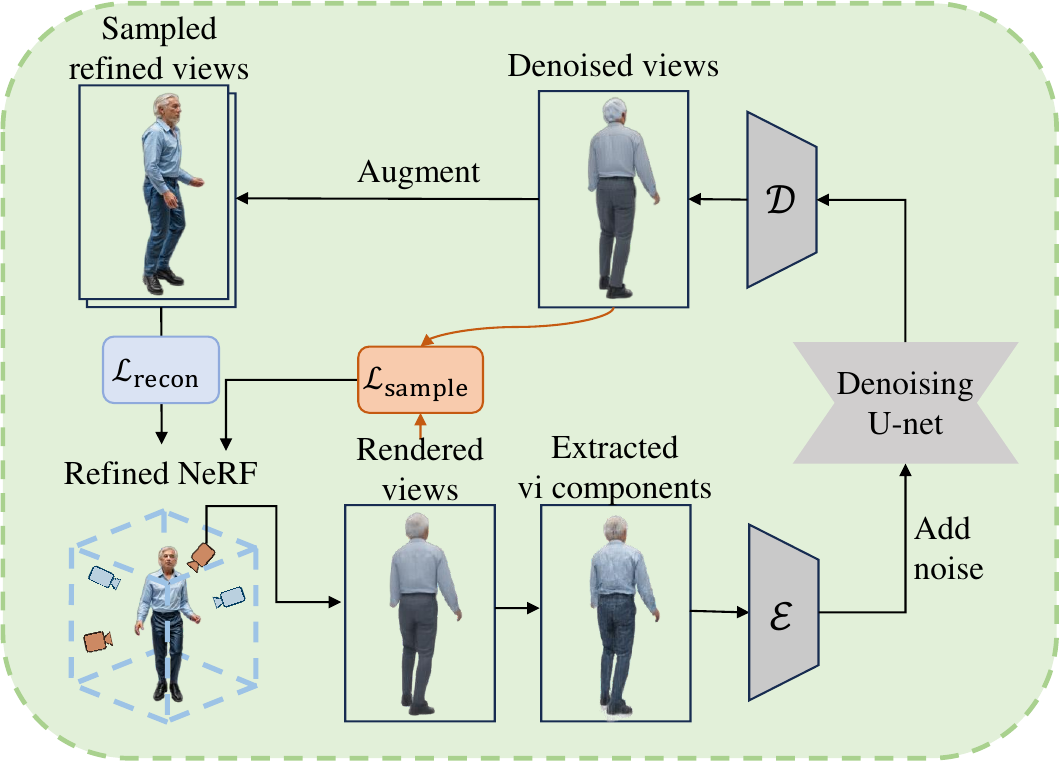}
    \caption{Progressive-Augmented Reconstruction.}
    \label{fig:par}
\end{subfigure}
\caption{\textbf{InceptionHuman overview} where all RGB images are generated synthetically. Given a combination of prompts (e.g., text prompt {\it an old white man with grey hair wearing blue shirt} and optionally others), we first generate a seed image with ControlNet and utilize a mesh estimation module to generate a series of coarse views. a) InceptionHuman introduces \emph{Iterative Pose-Aware Refinement} (IPAR), where the coarse views are used to reconstruct a radiance field, and we iteratively extract the view-independent components to produce refined views. b) Finally, InceptionHuman adopts the novel \emph{Progressive-Augmented Reconstruction} (PAR), which includes a convergent and 
view-independent regularization to match the consistency among the refined images and remove undesirable artifacts.} 
\label{fig:overview}
\vspace{-1em}
\end{figure}

After preprocessing, \ourwork~consists of two  stages: 
\textit{Iterative Pose-Aware Refinement} (IPAR) and \textit{Progressive-Augmented Reconstruction} (PAR). Figure~\ref{fig:overview} shows the technical overview of IPAR and PAR.



\begin{figure}[t]
    \centering
    \includegraphics[width=0.92\linewidth]{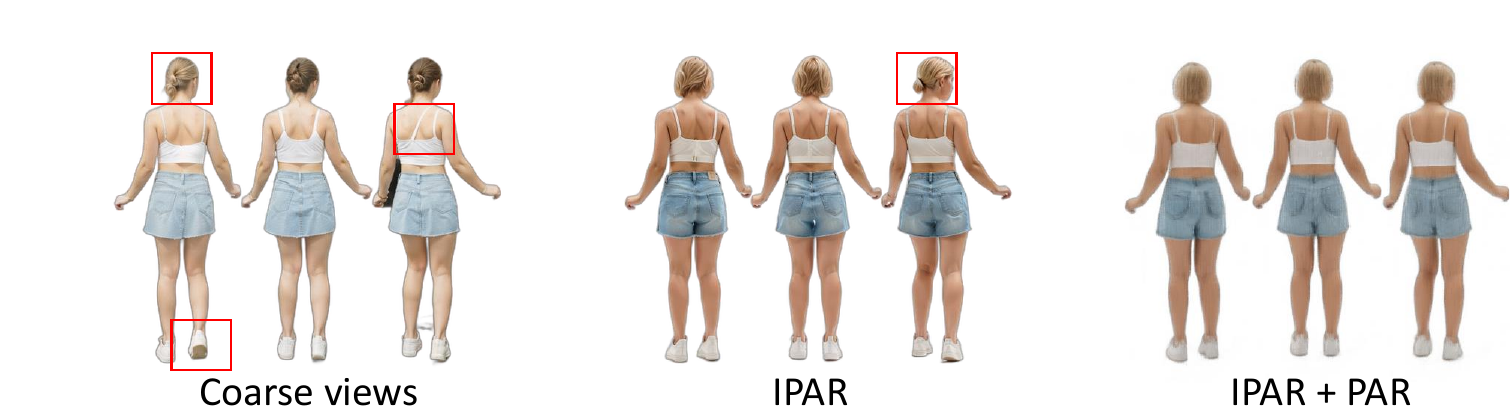}
    \caption{Coarse views generated from diffusion models naively suffer from serious 3D inconsistency. This issue can be solved by adopting IPAR and PAR.}
    \label{fig:progression}
    \vspace{-1em}
\end{figure}

\vspace{2mm}
\noindent{\textbf{Preprocessing.}}
In InceptionHuman, we first generate a number of synthetic views, referred as coarse views, based on the given prompts $\{P_0, P_1, \cdots, P_k\}$, where $P_0$ indicates the text control, and $P_1, P_2, ..., P_k$ represent additional optional image controls, i.e., depth, edges, pose, etc. We introduce a diffusion model $\bG_1$ that generates a 2D human image $\mathcal{I}_{\it{seed}}$ as follows:
\begin{equation}
    \mathcal{I}_{\it{seed}} = \bG_1\left(P_0, P_1, ..., P_k\right).
\end{equation}
This seed image $\mathcal{I}_{\it{seed}}$ provides the high-level features of our generated human prompted by the user. A coarse 3D geometry can then be generated from $\mathcal{I}_{\it{seed}}$. We leverage a single-view mesh prediction model~\cite{econ} to generate a 3D mesh $\mathcal{M}$ conditioned on $\mathcal{I}_{\it{seed}}$. Note that this coarse mesh serves only a 3D geometry proxy instead of the final geometry due to its crucial noise. We render $n$ views from the mesh $\mathcal{M}$, namely $m_1, m_2, ..., m_n$, and associate each rendered mesh view $m_i$ with 
$k$ control types, (\emph{e.g.,} edges, depth, etc) denoted as $f_1(m_i), f_2(m_i), ..., f_k(m_i)$. These mesh views are then be used as controls to generate coarse views from a pretrained diffusion model $\bG_2$:
\begin{equation}
    v_{i} = \bG_2\left(P_0, \mathcal{I}_{\it{seed}}, f_1(m_i), f_2(m_i), ..., f_k(m_i)\right)
\end{equation}
for $i=1, 2, ..., n$, where $v_i$ is the coarse view generated from $m_i$, associated with the camera pose of $m_i$. Note that the generated $v_i$'s are inconsistent and 3D-unaware due to the missing cross-view knowledge, as shown in Figure~\ref{fig:progression}.

\subsection{Iterative Pose-Aware Refinement (IPAR)} Built upon the assumption that rendered views with close camera poses share similar RGB values, IPAR is proposed to remove the inconsistency among coarse views $v_i$'s and generate 3D-aware refined views.  We iteratively add substantial details to the coarse views frame by frame with the assistance of diffusion models. In particular, we first use the 
Clean-NeRF~\cite{cleannerf} architecture, where colors are decomposed into view-independent component $\mathbf{c}_\text{vi}$ and view-dependent component 
$\mathbf{c}_\text{vd}$:
\begin{equation}
\label{eqn:blend}
    \mathbf{c} = \gamma\mathbf{c}_{\text{vi}} + (1-\gamma)\mathbf{c}_{\text{vd}},
\end{equation}
with $\gamma$ as a weight factor. We reconstruct the coarse views from preprocessing with reconstruction loss $\mathcal{L}_{\text{recon}}$, consisting of a photorealistic loss $\mathcal{L}_{\text{pho}}$, a view-independent regularization $\mathcal{L}_{\text{vi}}$, and a view-dependent regularization $\mathcal{L}_{\text{vd}}$
(detailed derivation in~\cite{cleannerf}):
\begin{equation}
    \label{consistent}\mathcal{L}_{\text{recon}}=\mathcal{L}_{\text{pho}}+\sum_{\mathbf{x}}\left(\mathcal{L}_{\text{vi}}+\mathcal{L}_{\text{vd}}\right),
\end{equation}
referring the resulting radiance field as coarse NeRF $\mathcal{F}$. We select this architecture over vanilla NeRF model to remove the inconsistency among coarse views, where Figure~\ref{fig:vi_component} in the ablation study shows that vi components successfully preserve more high-level features.

We leverage a diffusion model $\bG_3$ in IPAR to denoise the view-independent components in coarse NeRF $\mathcal{F}$. Suppose we render $n$ view-independent frames $v^{\text{ind}}_1,v^{\text{ind}}_2,...,v^{\text{ind}}_n$ from $\mathcal{F}$ with the strategy: 
\begin{equation}
\label{cvi_rend}
    \hat{\mathbf{C}}_\text{vi} = \sum_{k=1}^K \hat{T}(t_k)\alpha(\sigma(t_k)\delta_k)\mathbf{c_{\text{vi}}}(t_k), 
\end{equation}
where $\hat{T}(t_k)=\operatorname{exp}\left(-\sum_{k^{'}=1}^{k-1}\sigma(t_k)\delta(t_k)\right)$, $\alpha \left({x}\right) = 1-\exp(-x)$, $\delta_k = t_{k+1} - t_k$, $t_k$ is the $k$-th sampled point along the ray, and $\sigma$ is the density. Different from $\bG_1$ and $\bG_2$, the diffusion model $\bG_3$ uses the \textit{img2img} module in Latent Diffusion Model~\cite{stablediffusion} to refine $v^{\text{ind}}_i$ into $v'_i$ with the same camera pose. This process can be written as
\begin{equation}
\label{refine}
    v'_{i} = \bG_3\left(v^{\text{ind}}_{i},v^{\text{ind}}_{i-1},P_0, \mathcal{I}_{\it{seed}}, f_1(m_i), f_2(m_i), ..., f_k(m_i)\right).
\end{equation}
Instead of synthesizing all views simultaneously, we propose to iteratively refine the views and finetune the coarse NeRF. This strategy imposes the continuous consistency between the rendered views for refined images. 
Algorithm~\ref{alg:refine} shows the overall process of IPAR. In each iteration, the algorithm refines a small number $c$ of independent views rendered from $\mathcal{F}$ and adds it to the image set $\mathcal{S}$ for further finetuning. 
Different from the Iterative Dataset Update (Iterative DU) in Instruct-NeRF2NeRF~\cite{instructnerf2023} that iteratively updates images until they converge, we update images in a ``segment-by-segment'' manner, and each view is updated only once. After this stage, IPAR enhances the inter-view consistency 
and adds more high-frequency view-independent details to the synthetic human images.

\begin{algorithm}[t]
\caption{Iterative Pose-Aware Refinement (IPAR)}\label{alg:refine}
\textbf{Input}: Coarse NeRF $\mathcal{F}$.\\
\textbf{Parameters}: \# refined views per iteration $c$.
\begin{algorithmic}[1]
    \State $\mathcal{S}\gets \{\}$.
    \For{$i=0,1,2,\cdots,n/c$}
    \State Render $\{v^{\text{ind}}_{ic+1},v^{\text{ind}}_{ic+2},\cdots,v^{\text{ind}}_{ic+c}\}$ from $\mathcal{F}$.
    \For{$j=1,2,\cdots,c$}
    \State Generate $v'_{ic+j}$ by Equation~\ref{refine}.
    \State $S\gets S\cup\{v'_{ic+j}\}$.
    \EndFor
    \State Finetune $\mathcal{F}$ with  $\mathcal{S}$ by Equation~\ref{consistent}.
    \EndFor
    \State Return $\mathcal{S}$.
\end{algorithmic}
\end{algorithm}

\subsection{Progressive-Augmented Reconstruction (PAR)} This stage further enhances the reconstruction quality and removes undesirable artifacts. 
We introduce a pretrained diffusion prior composed of encoder $\mathcal{E}$, denoising U-net $\bG_4$, and decoder $\mathcal{D}$. Recall that the IPAR assumes that views with close camera poses have similar RGB values. Hence, IPAR captures consistency for closed views. In contrast, PAR has the following assumption: \textit{``The inconsistency in the synthetic dataset is less noticeable when views are sparse.''} This assumption is straightforward. It captures multi-view consistency across sparse view reconstruction. Different from IPAR, PAR considers consistency across far view points. Based on this assumption, PAR samples a sparse set of refined views $v'_i$'s in $\mathcal{S}$, denoted as $\mathcal{S}^{\text{sub}}$, and progressively augment $\mathcal{S}^{\text{sub}}$ with 3D-consistent views. We use $\mathbf{C}$ and $\mathbf{C}'$ to respectively denote the sets of camera poses of views in $\mathcal{S}^{\text{sub}}$ and $\mathcal{S} - \mathcal{S}^{\text{sub}}$. In each back-propagation step, PAR randomly samples and renders $d$ view-independent components with Equation~\ref{cvi_rend} from $\mathbf{C}'$ if it is not empty, denoted as $w^{\text{ind}}_1, w^{\text{ind}}_2, ..., w^{\text{ind}}_d$, with respect to the full-colored, \ie $(\text{vi}+\text{vd})$, views $w_1, w_2, ..., w_d$. These vi components are sent to diffusion prior to regularize the NeRF reconstruction. Specifically, for each vi component $w^{\text{ind}}_i$, we generate a ``refined'' vi component with the following operation:
\begin{algorithm}[t]
\caption{Progressive-Augmented Reconstruction (PAR)}\label{alg:par}
\textbf{Input}: Refined views $\mathcal{S}$.\\
\textbf{Parameters}: \# sampled novel views $d$, \# steps in each iteration $n$, range of denoising strength $\hat{t}_{\text{max}}$ and $\hat{t}_{\text{min}}$.
\begin{algorithmic}[1]
    \State $\mathcal{S}^{\text{sub}}\gets\text{sample}(\mathcal{S})$.
    \For{each iteration}
    \State $\mathbf{C}\gets \text{Poses of }\mathcal{S}^{\text{sub}} \text{;  }\mathbf{C}'\gets \text{Poses of }(\mathcal{S}-\mathcal{S}^{\text{sub}})$.
    \State Initiate radiance field $\mathcal{F}$.
    \For{each back-propagating step $s$}
    \State $t_{\text{max}}\gets \hat{t}_{\text{max}}-\frac{s}{n}(\hat{t}_{\text{max}}-\hat{t}_{\text{min}})$.
    \State Randomly sample and render $d$ view pairs $\{(w_i,w^{\text{ind}}_i)\}$ from  $\mathbf{C}'$.
    \State Sample $t\in(0,t_{\text{max}}]$.
    \State Generate views $\{w'_i\}$ by Equation~\ref{denoise}.
    \State Back-propagate $\mathcal{F}$ with $\mathcal{L}_{\text{recon}}+\mathcal{L}_{\text{sample}}$ by Equation~\ref{consistent} and~\ref{sample_loss}.
    \EndFor
    \State Sample views from $\mathbf{C}'$ and generate $W=\{w'_i\}$ by Equation~\ref{denoise}.
    \State $\mathcal{S}^{\text{sub}}\gets\mathcal{S}^{\text{sub}}+W$
    \EndFor
    \State Return $\mathcal{F}$.
\end{algorithmic}
\end{algorithm}
\begin{equation}
    \label{denoise}
    w'_i = \mathcal{D}\left(\bG_4^t\left(\mathcal{E}\left(w^{\text{ind}}_i\right)+t\epsilon\right)\right),
\end{equation}
where $\epsilon$ is a random Gaussian noise, $t$ is a random number sampled from $(0, t_{\text{max}}]$ with a hyper-parameter $t_{\text{max}}$, and $\bG_4^t$ denotes using $\bG_4$ as backbone with denoising strength $t$. For simplicity, we ignore the latent controls of $\bG_4$ in Equation~\ref{denoise}. Similar to IPAR, we also use the input text prompt and mesh image $f(m_j)$ to guide this denoising process. We then compute the view-independent regularizer with the definition:
\begin{equation}
    \label{sample_loss}
    \mathcal{L}_{\text{sample}} = \mathbb{E}_t\left(\mathcal{L}_p(w'_i, w_i)+t\lVert w'_i-w_i\rVert_1\right),
\end{equation}
where $\mathcal{L}_p$ refers to the perceptual distance LPIPS~\cite{zhang2018unreasonable}. 

PAR is conducted progressively in iterations. We illustrate the detailed process in Algorithm~\ref{alg:par}. In each iteration, we use Equation~\ref{consistent} and~\ref{sample_loss} to reconstruct a radiance field. This radiance field is next used to augment $\mathcal{S}^{\text{sub}}$ with Equation~\ref{denoise}, where we render a sampled set of view-independent components from $\mathbf{C}'$ and add the denoised images to $\mathcal{S}^{\text{sub}}$. Note that the camera poses in $\mathbf{C}$ and $\mathbf{C}'$ are simultaneously updated with respect to $\mathcal{S}^{\text{sub}}$. This progression terminates when $\mathbf{C}'$ is empty, where we obtain a denoised dataset $\mathcal{S}^{\text{sub}}$ with the same cardinality and camera poses as $\mathcal{S}$. This dataset $\mathcal{S}^{\text{sub}}$ is finally used to reconstruct the 3D human of our InceptionHuman with Equation~\ref{consistent} and~\ref{sample_loss}. Note that the sample regularizer samples vi components in $\mathbf{C}$ in this final iteration, as $\mathbf{C}'$ is empty. $t_{\text{max}}$ is linearly decayed from $\hat{t}_{\text{max}}$ to $\hat{t}_{\text{min}}$ within the steps of each progressive iteration. 

\section{Experiments}\label{sec:experiment}
This section first summarizes the implementation details in section~\ref{implement} and presents qualitative and quantitative comparison of InceptionHuman with most recent state-of-the-arts in text-to-3D human generation in Section~\ref{text_to_3D}. Then, we present multi-control generation beyond text in Section~\ref{multicontrol} to demonstrate how users can easily condition and control \ourwork's prompt-to-NeRF generation using additional handy input modalities. Next we conduct ablation studies in Section~\ref{ablation} to validate the efficacy of individual components of \ourwork, investigating the effect of the view-dependent components, IPAR and PAR.

\subsection{Implementation details}\label{implement}
Here we only illustrate the model structure and hyper-parameters of IPAR and PAR. For other details such as the training recipes, please refer to the appendix.

\vspace{2mm}
\noindent{\textbf{Diffusion models.}} In this paper, all diffusion models are not finetuned on any task-specific datasets. We use the open-source ``\textit{epiCRealism}'' of Stable Diffusion (SD) backbone, appending prompt-corresponding ControlNet models to build our diffusion network. In specific, $\bG_1$ includes $k$ ControlNets depending on the size of sequence of input prompts. For other models, the following strategies are adopted: 1) Style-based model is used for seed image control; 2) depth and edge controls of mesh rendered image are included for $\bG_2, \bG_3$, and $\bG_4$; 3) soft-edge control of the neighbor view is included in $\bG_3$.

\vspace{2mm}
\noindent{\textbf{Coarse mesh.}} We utilize the golden ECON~\cite{econ} module for mesh prior estimation. After generating the mesh, we use Blender~\cite{blender} to build the rendering viewpoints and their corresponding camera poses. This allows for rendering multiple 2D images and depth maps of the mesh object, which are later used for synthetic view generation.

\vspace{2mm}
\noindent{\textbf{Hyper-parameters.}} In IPAR, we set the step size hyper-parameter $c$ to 5 with 200 images in total, and the number of finetuning steps in each iteration linearly increase from 1200 to 2000 throughout the refining process. In PAR, we first sample 20 images from the refined images and use 4 progressive iteration for augmentation. Each iteration is trained with $n=5000$ steps, and the number of sampled vi components $d$ is set 4.The maximal denoising strength $\hat{t}_{\text{max}}$ decreases from $1.0$ to $0.4$ over each iteration, and the minimal denoising strength $\hat{t}_{\text{min}}$ is set 0.1. Due to the limitation of our computing resources, we randomly sample a chunk in each denoised view in Equation~\ref{sample_loss} for back-propagation instead of using the entire image. 

\subsection{Text-to-3D comparisons}\label{text_to_3D} 
We compare our work with two state-of-the-art text-guided and realistic-styled 3D human generative modules, DreamHuman~\cite{dreamhuman} and HumanNorm~\cite{humannorm}, as these two methods have shown significant improvement over previous works in Figure~\ref{fig:sota}. While the goals are the same, i.e., generating 3D human, they are different methods:
\begin{figure}[t!]
\centering
\resizebox{.9\linewidth}{!}{%
\begin{tabular}{cccc}
    \large{Text description} & \large{DreamHuman} & \large{HumanNorm} & \large{InceptionHuman (Ours)}\\\toprule
    \raisebox{-\totalheight}{\mline{\large{\textit{``a woman wearing a short jean skirt and a cropped top''}}}} & \raisebox{-\totalheight}{\includegraphics[height=70mm]{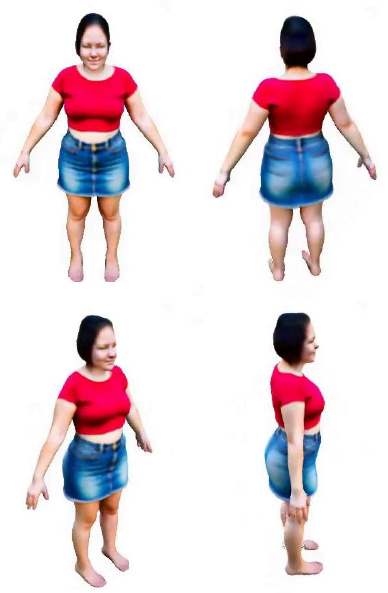}} & \raisebox{-\totalheight}{\includegraphics[height=70mm]{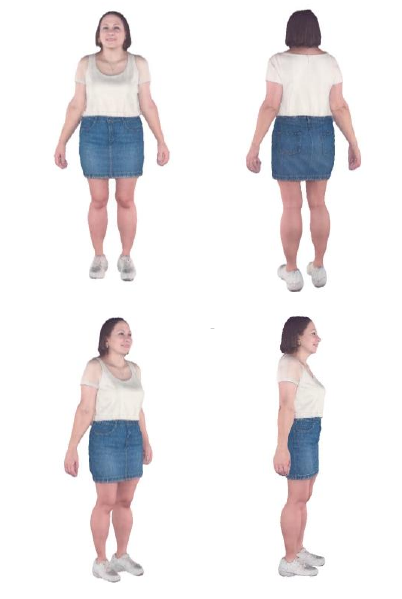}} & \raisebox{-\totalheight}{\includegraphics[width=.4\columnwidth]{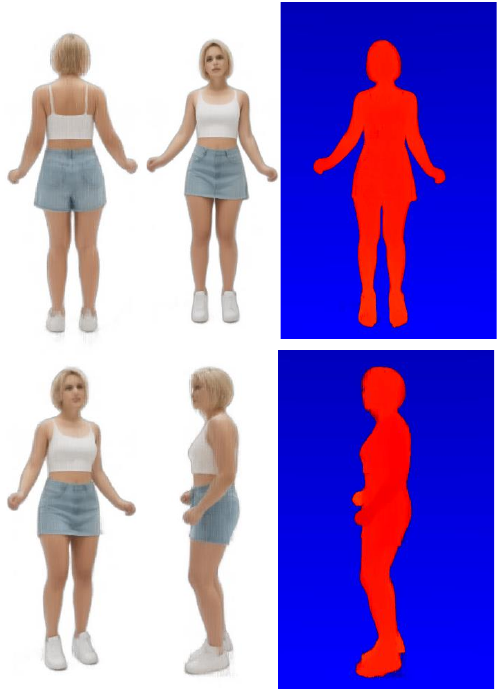}}\\\midrule
    
    \raisebox{-\totalheight}{\mline{\large{\textit{``a man wearing a jean jacket and jean trousers''}}}}& \raisebox{-\totalheight}{\includegraphics[height=70mm]{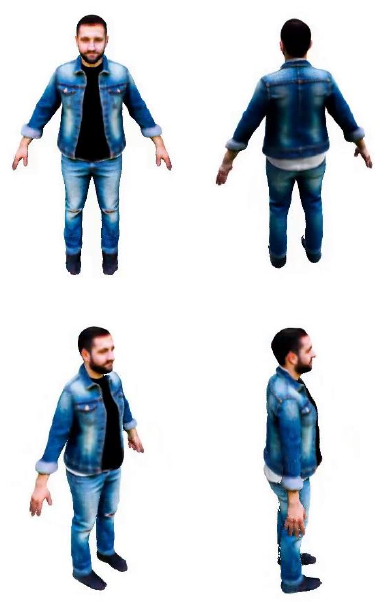}} & \raisebox{-\totalheight}{\includegraphics[height=70mm]{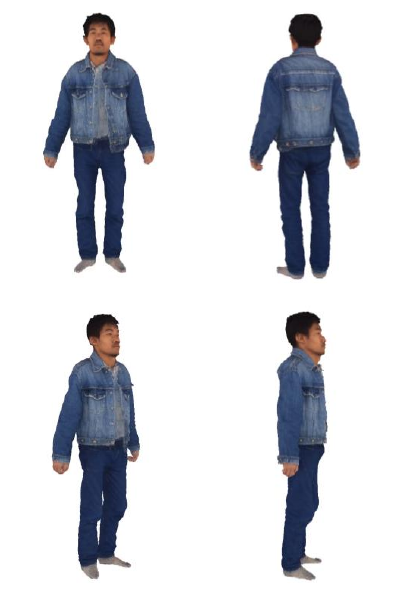}} & \raisebox{-\totalheight}{\includegraphics[width=.4\columnwidth]{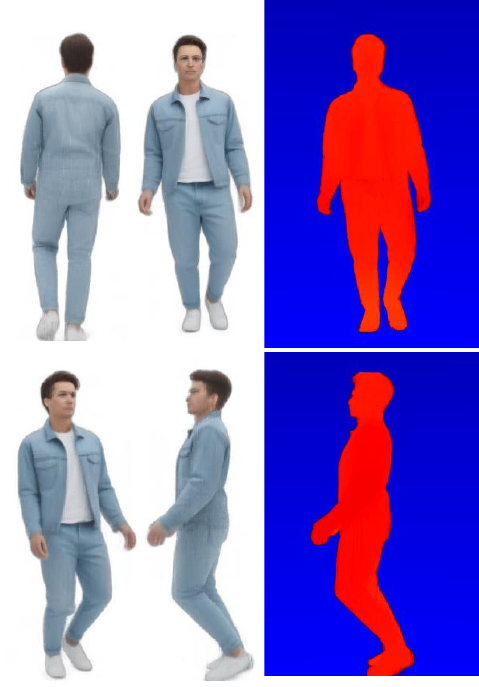}}\\\midrule
    
    \raisebox{-\totalheight}{\mline{\large{\textit{``a man wearing a white tanktop and shorts''}}}}& \raisebox{-\totalheight}{\includegraphics[height=70mm]{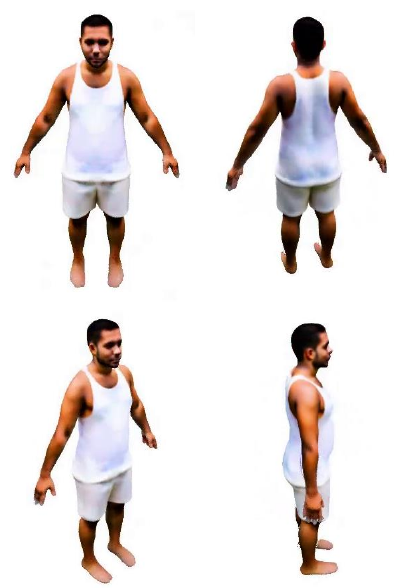}} & \raisebox{-\totalheight}{\includegraphics[height=70mm]{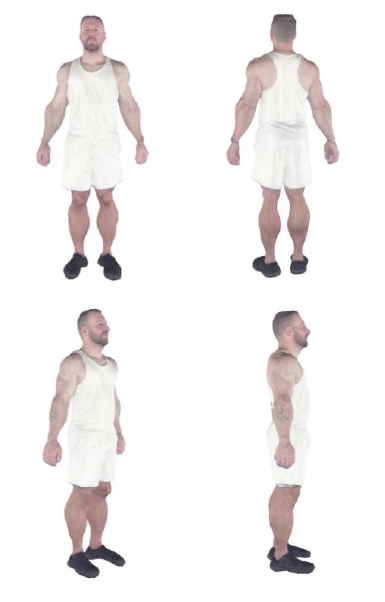}} & \raisebox{-\totalheight}{\includegraphics[width=.4\columnwidth]{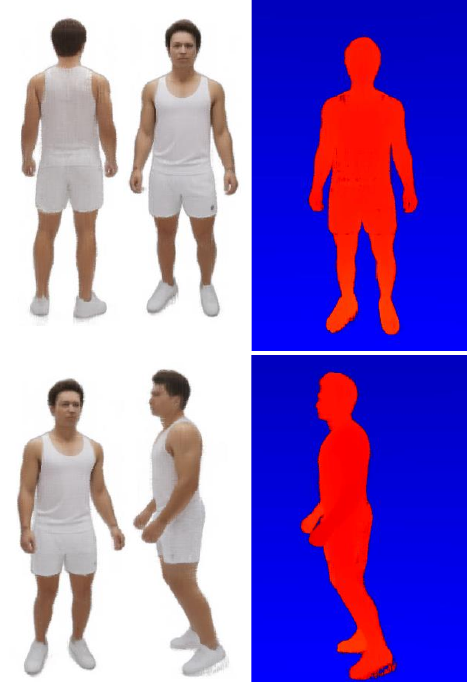}}\\

\end{tabular}
}
\caption{\textbf{Text-to-3D qualitative comparison.} Zoom in for details of InceptionHuman's results and the corresponding depth maps estimated from NeRF. See also the supplemental video.}
\label{tab:qualitative}
\vspace{-1.3em}
\end{figure}
\begin{itemize}
    \item \textbf{DreamHuman}: A NeRF-based method with text inputs to generate animatable 3D humans/avatars. 
    \item \textbf{HumanNorm}: A conventional approach for 3D human generation employing mesh for 3D geometry and RGB image for texture mapping, leveraging recent stable diffusion model for high-quality mesh and texture generation. 
    \item \textbf{\ourwork}: A NeRF-based approach incorporating  ControlNet for high-quality static human generation with multi-modal input prompts (e.g., texts, poses).
\end{itemize}

\vspace{2mm}
\noindent{\textbf{Qualitative comparisons.}} Given identical text descriptions, Figure~\ref{tab:qualitative} shows the qualitative comparison of \ourwork~with 
DreamHuman~\cite{dreamhuman} and
HumanNorm~\cite{humannorm}. While HumanNorm improves the quality by decoupling 3D human into the geometry and texture generation, the issue of unnatural colors can still be observed, and some of their results are not aligned with the given text prompts, \eg, the ``crop top'' example in Figure~\ref{tab:qualitative}. On the other hand, \ourwork~naturally overcomes these issues. The results also show that our work can generate more photorealistic colors and characters with less artifacts. 

\vspace{2mm}
\noindent{\textbf{Quantitative comparisons.}} We compare the quantitative quality of our proposed InceptionHuman using two metrics. First, we use the commonly adopted CLIP to measure the average semantic similarity among all pairs of views. In specific, for a set of $n$ rendered views $\mathcal{S}=\{v_1,v_2,...,v_n\}$, the average semantic similarity can be represented as $\frac{1}{{\binom{n}{2}}}\sum_{i<j}Sc\left(\Phi\left(v_i\right),\Phi\left(v_j\right)\right)$, where $Sc(\cdot)$ is cosine similarity and $\Phi(\cdot)$ is  CLIP encoding. This is used to measure the consistency of high-level features, such as distinct clothes colors from different views. Similar definition of semantic similarity can be found in DietNeRF~\cite{dietnerf}. Table~\ref{tab:quantitative} shows that our work achieves higher semantic similarity. We explain this from the texture aspect; specifically, InceptionHuman generates more natural/photorealisitic colors that reduce the undesirable noises in CLIP embeddings.
\begin{figure}
\centering
\resizebox{.97\linewidth}{!}{%
\begin{tabular}{cc}
    {Multi-control Prompts}\tab[1cm] & \tab[1cm]{3D generation results}\\\midrule
    
    \raisebox{-\totalheight}{\includegraphics[height=21mm]{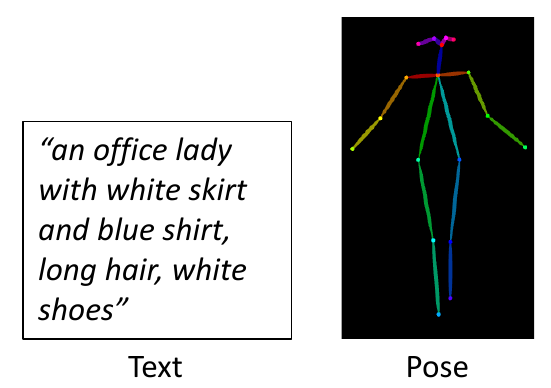}}& \raisebox{-\totalheight}{\includegraphics[width=.6\columnwidth]{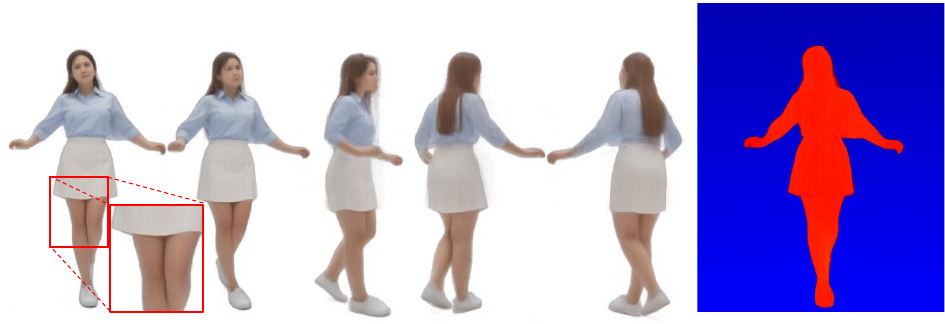}}\\\midrule
    \raisebox{-\totalheight}{\includegraphics[height=21mm]{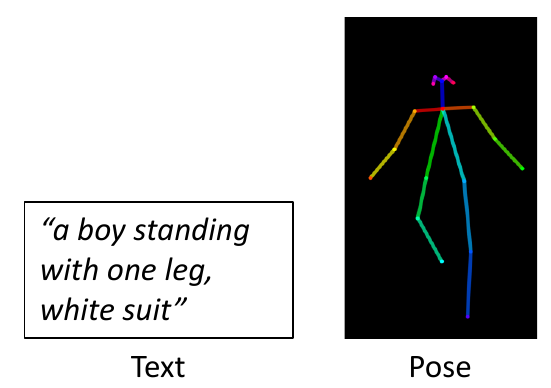}}&\raisebox{-\totalheight}{\includegraphics[width=.6\columnwidth]{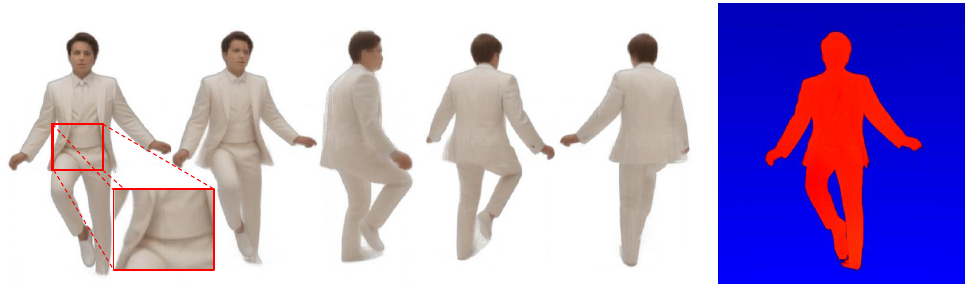}} \\\midrule

    \raisebox{-\totalheight}{\includegraphics[height=21mm]{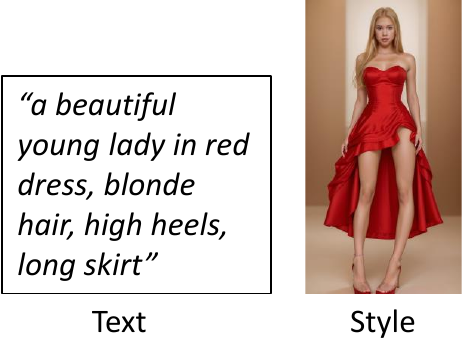}}& \raisebox{-\totalheight}{\includegraphics[width=.6\columnwidth]{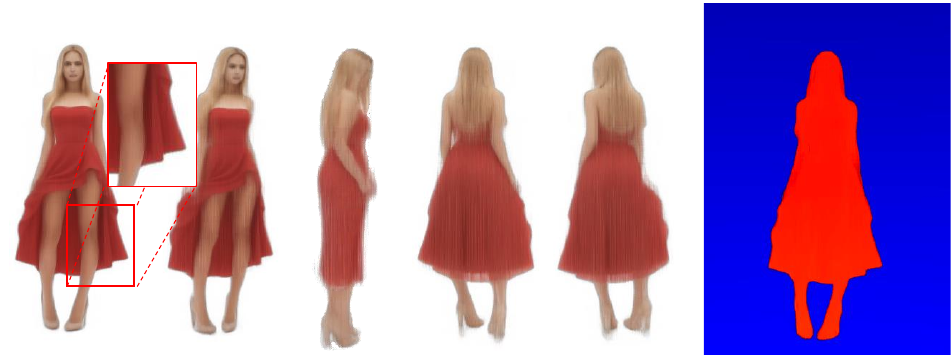}} \\\midrule

    \raisebox{-\totalheight}{\includegraphics[height=21mm]{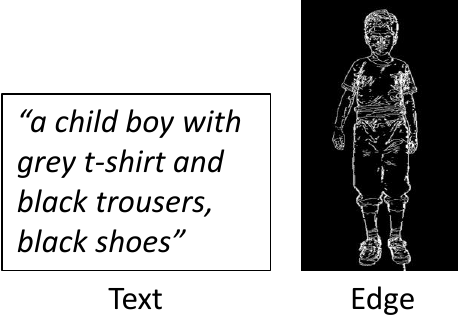}} &\raisebox{-\totalheight}{\includegraphics[width=.6\columnwidth]{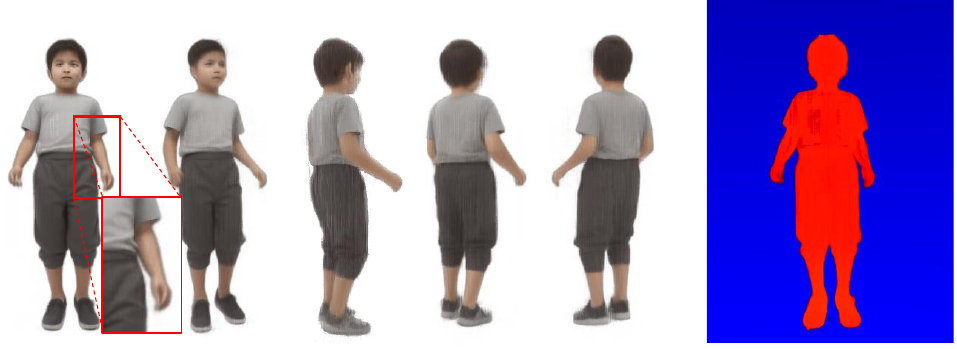}} \\\midrule

    \raisebox{-\totalheight}{\includegraphics[height=21mm]{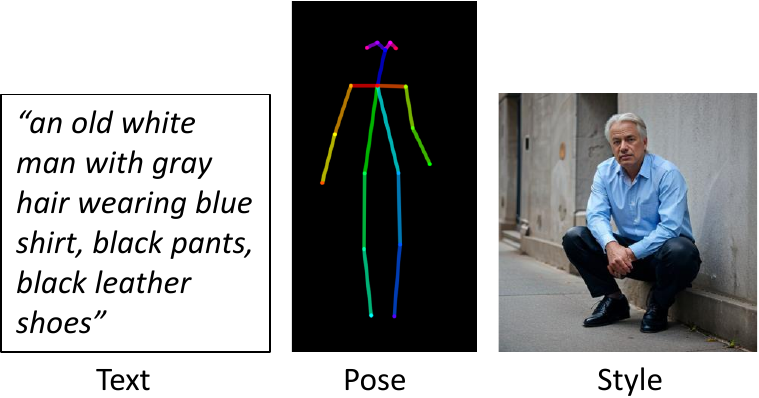}}& \raisebox{-\totalheight}{\includegraphics[width=.6\columnwidth]{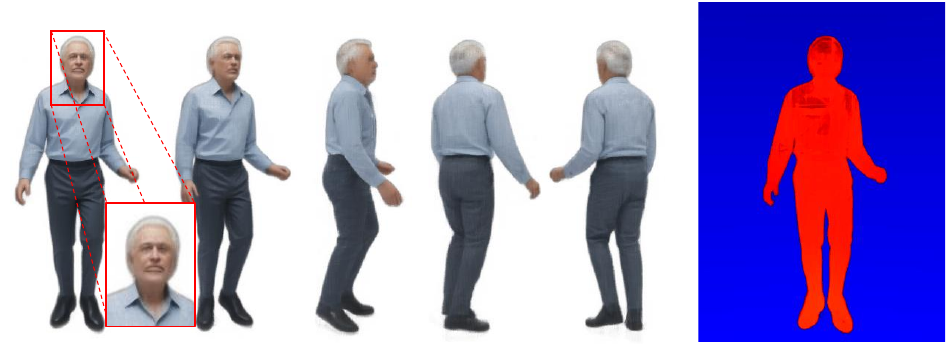}} \\\midrule
    
    \raisebox{-\totalheight}{\includegraphics[height=21mm]{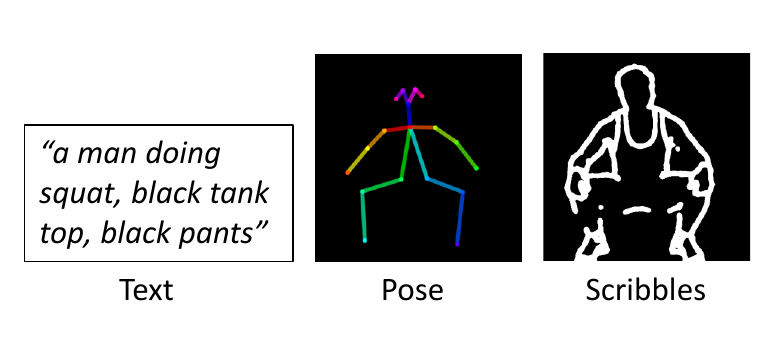}} &\raisebox{-\totalheight}{\includegraphics[width=.6\columnwidth]{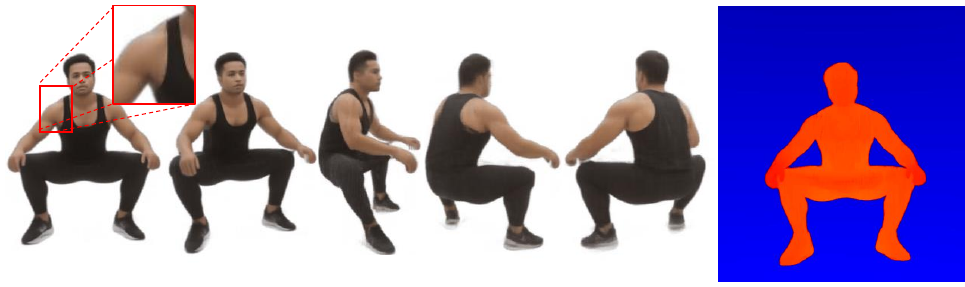}} \\
\end{tabular}
}
\caption{\textbf{Qualitative results with multi-control prompts.} Zoom in for details. See also the supplemental materials for reference.}
\label{tab:results}
\end{figure}

\begin{figure}[t!]
    \centering
    \begin{subfigure}{0.32\linewidth}
    \includegraphics[width=\textwidth]{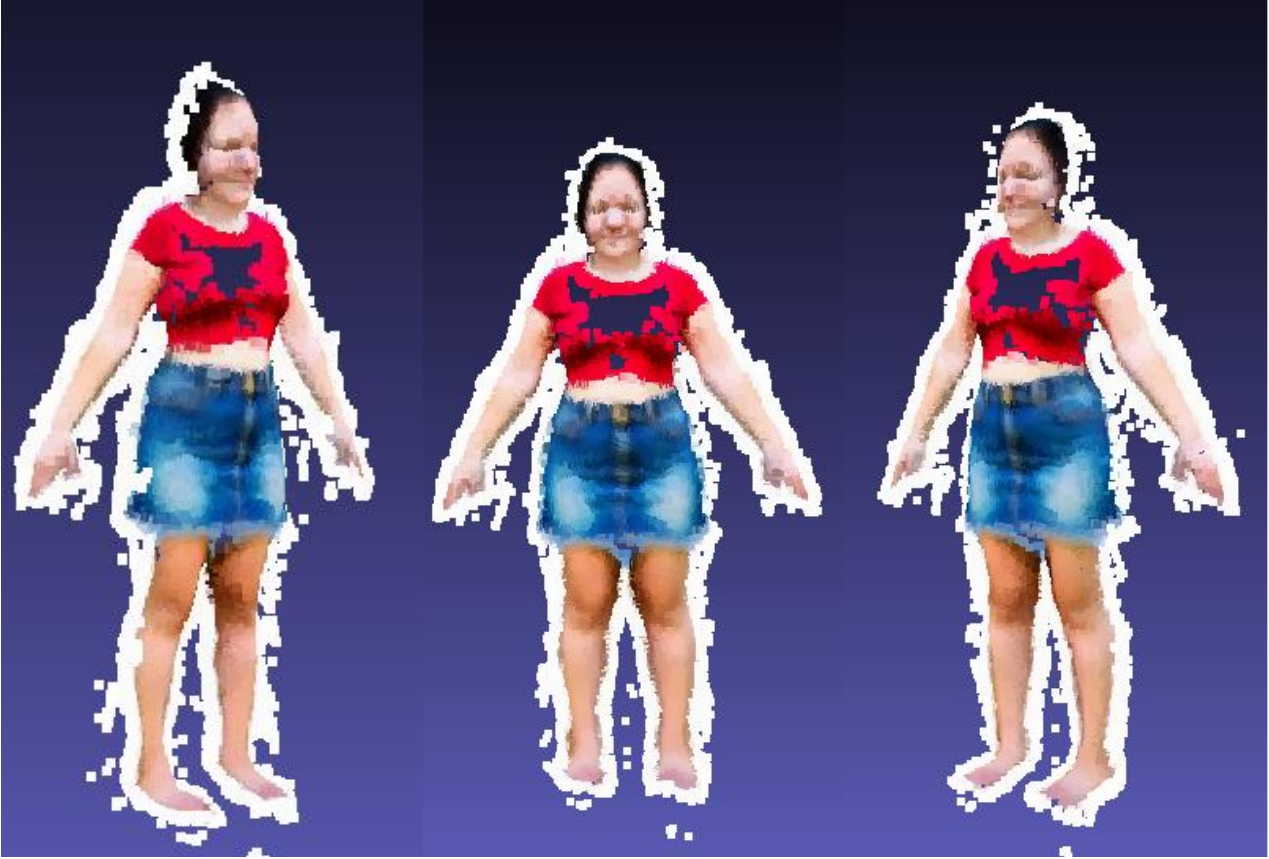}
    \caption{DreamHuman}
    \label{fig:dreamhuman_colmap}
\end{subfigure}
\hfill
\begin{subfigure}{0.32\linewidth}
    \includegraphics[width=\textwidth]{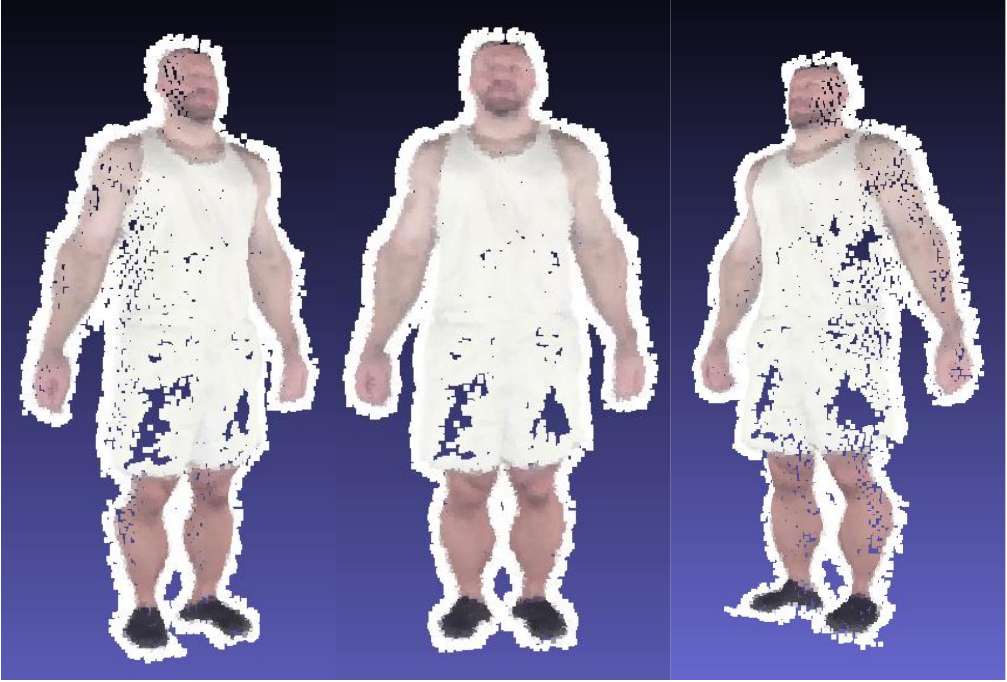}
    \caption{HumanNorm}
    \label{fig:humannorm_colmap}
\end{subfigure}
\hfill
\begin{subfigure}{0.32\linewidth}
    \includegraphics[width=\textwidth]{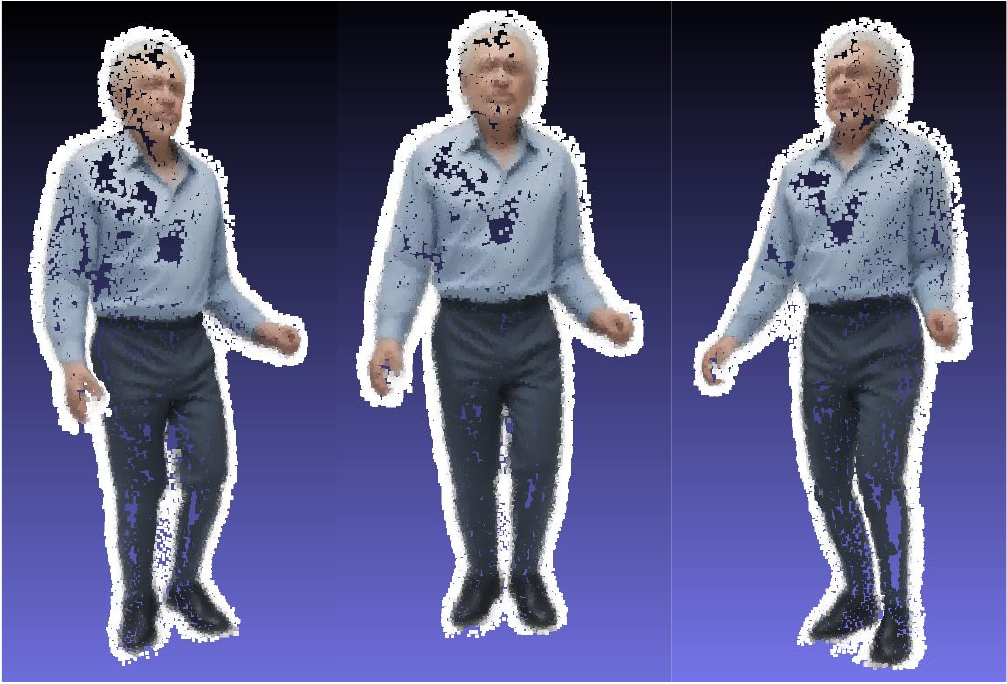}
    \caption{InceptionHuman}
    \label{fig:inceptionhuman_colmap}
\end{subfigure}
\caption{\textbf{COLMAP multiview quasi-dense reconstruction} with 10 views.}
\vspace{-1.5em}
\label{fig:colmap}
\end{figure}

Additionally, we sample 10 near frontal views in the generated 3D humans and employ COLMAP~\cite{COLMAP}, a Structure-from-Motion (SfM) and Multi-View Stereo (MVS) pipeline, to measure the mean reprojection error. For each of three generation methods, this measurement is conducted over three example characters. Table~\ref{tab:quantitative} shows that our InceptionHuman has lower reprojection error than DreamHuman and HumanNorm. Moreover, our synthetic images can be employed in conventional multiview stereo reconstruction: Figure~\ref{fig:colmap} presents the quasi-dense multiview stereo reconstruction result with 10 near front views {\em synthetically} generated for visualization, which shows that our work generates 3D-consistent images.
\begin{table}
    \centering
      \resizebox{.8\columnwidth}{!}{    \begin{tabular}{c|ccc}
    \toprule
    & DreamHuman & HumanNorm & \ourwork \\\midrule
Avg. semantic similarity $\uparrow$ & 0.8883 & 0.9106 & \bf0.9240 \\
Mean reprojection error $\downarrow$&0.8874&0.8730&\bf0.7695\\
\bottomrule
    \end{tabular}
    }
    \caption{\textbf{Quantitative comparisons with baselines.} We use CLIP to measure the semantic similarity between different views and use COLMAP to measure the reprojection error of front views.}
    \label{tab:quantitative}
    \vspace{-1.3em}
\end{table}

\subsection{Multi-control generation}\label{multicontrol}
One of our main strengths beyond existing 3D human generation is controllable generation with multiple prompts. Figure~\ref{tab:results}  demonstrates the results of our InceptionHuman with different types of controls, allowing a user to easily generate 3D human of a given apparel style and/or pose in addition to text prompt.

\subsection{Ablation studies}\label{ablation}
In addition to Figure~\ref{fig:progression}, which shows the effectiveness of our proposed IPAR and PAR, we in this subsection conduct extra experiments to verify the methodology.

\begin{figure}[t!]
    \centering
    \includegraphics[width=.8\linewidth]{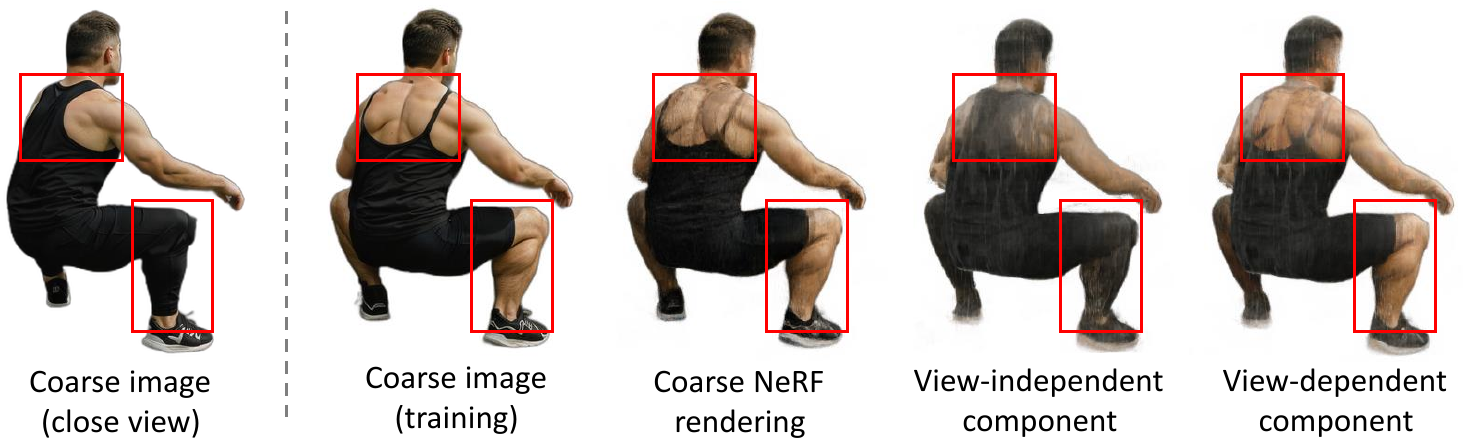}
    \caption{\textbf{Effect of view-independent (vi) components.} Compared to naive rendering and view-dependent (vd) components, vi components preserve more high-level features in the inconsistent coarse images. To remove the inconsistency in coarse images, we only retain the view-independent (vi) components as inputs for the refining diffusion models.
    }
    \label{fig:vi_component}
    \vspace{-1em}
\end{figure}

\vspace{2mm}
\noindent{\textbf{Discarding vd components.}} In IPAR and PAR, we proposed to discard the view-dependent (vd) components during the inference of diffusion models. This is because vd components contains high-frequency rendering values, which are undesirable for inconsistent views and can cause 3D artifacts. In Figure~\ref{fig:vi_component}, we show that, when the coarse image has inconsistent RGB values against its close views, the vi component can preserve the 3D consistency, while the vanilla NeRF rendering and vd components cannot.

\begin{figure}[t!]
    \centering
    \begin{subfigure}{0.52\linewidth}
    \centering
    \includegraphics[width=\linewidth]{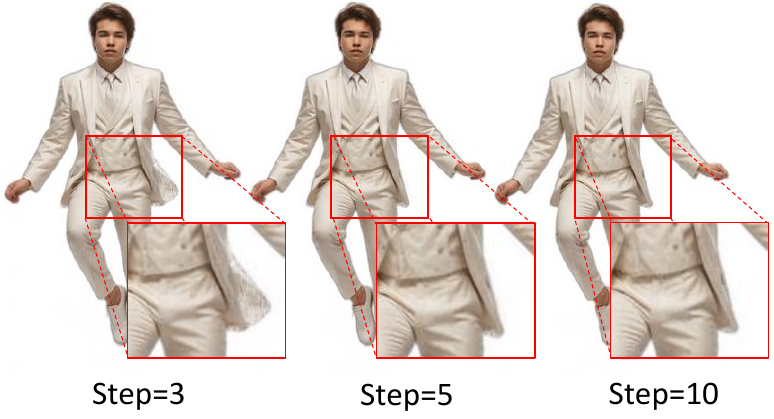}
    \caption{IPAR}
    \label{fig:IPAR}
\end{subfigure}
\hfill
\begin{subfigure}{0.46\linewidth}
\centering
    \includegraphics[width=\linewidth]{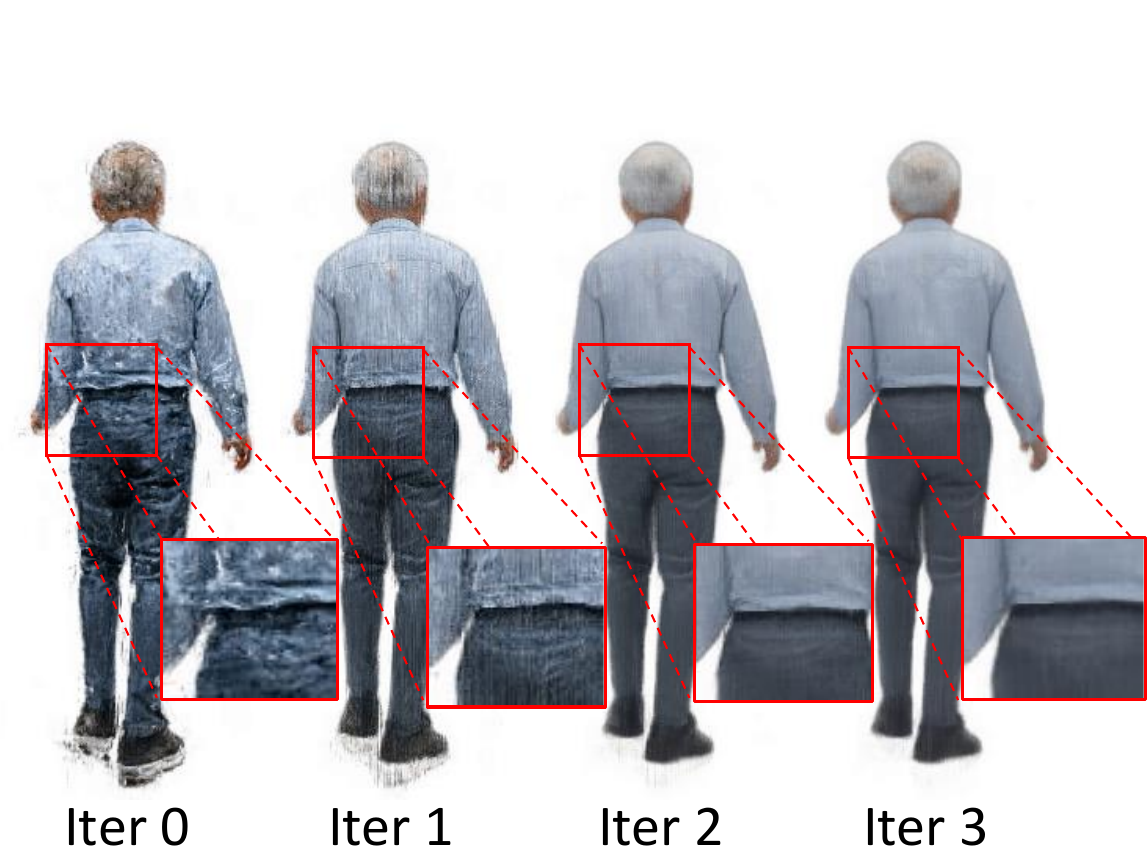}
    \caption{PAR}
    \label{fig:PAR}
\end{subfigure}
\caption{\textbf{Ablation studies.} IPAR and PAR progressively enhance the quality of synthetic views.}
\vspace{-1.2em}
\end{figure}

\vspace{2mm}
\noindent{\textbf{IPAR step size.}} We evaluate the IPAR quality with different step sizes, \ie, hyper-parameter $c$, by reconstructing radiance fields and rendering the corresponding views. We observe that the network tends to forget the geometry of coarse NeRF with small values of $c$ due to many iterations, and large values of $c$ cannot alleviate the inconsistency effectively. Overall, Figure~\ref{fig:IPAR} shows a recommended value in $[5,10]$.

\vspace{2mm}
\noindent{\textbf{PAR iteration.}} Recall that PAR starts by sampling a subset from PAR's outputs and progressively augment the dataset. In each iteration, we reconstruct the radiance field from scratch with $5000$ back-propagation steps. Figure~\ref{fig:PAR} shows that PAR successfully augments 3D-consistent images and enhances the quality of NeRF reconstruction with more iterations.

\section{Discussion}\label{sec:discussion}
While \ourwork~has presented promising results on generated photorealistic 3D human with NeRF, we would like to discuss the potential future development of our work. Previous works of text-to-3D human generation, such as TADA~\cite{tada}, DreamHuman~\cite{dreamhuman} and AvatarCLIP~\cite{avatarclip}, have demonstrated the applications on animatable 3D humans, where they leverage a deformable NeRF space for reconstruction. While our work can be controlled with human pose/skeleton like \cite{cao2023dreamavatar,zhang2023avatarverse}, the temporal coherence for animatable humans remains unexplored. Our work focuses on the progressive strategies to remove inconsistency among synthetic images. Possible methods to animate our results include: 1) generating temporal-coherent static 3D human per frame with continuous skeleton controls. 2) utilizing image-based animation modules such as AnimateDiff~\cite{guo2023animatediff} and AnimateAnyone~\cite{hu2023animateanyone}. 

\section{Conclusion}\label{sec:conclusion}
This paper proposes \ourwork, a novel end-to-end Prompt-to-NeRF framework that generates high-quality 3D human NeRF with a multi-modal array of guiding prompts, including a text description with additional controls such as mesh, pose and style. We capitalize on the state-of-the-art 2D controllable diffusion model with Neural Radiance Field (NeRF), and employ a two-stage NeRF reconstruction approach with \textit{Iterative Pose-Aware Refinement} (IPAR) and \textit{Progressive-Augmented Reconstruction} (PAR) to ensure the consistency among synthetic images. Extensive experiments have shown that our InceptionHuman outperforms the state-of-the-art baselines in quality, and substantially expand the applicability of 3D human generation for casual users in usability with its multi-control generation.

\newpage
\noindent{\Large\textbf{Appendix}}
\renewcommand\thesection{\Alph{section}}
\setcounter{section}{0}
\appendix
\section{Concurrent works}
While we have compared our InceptionHuman with DreamHuman~\cite{dreamhuman} (NeurIPS'23) and HumanNorm~\cite{humannorm} (CVPR'24), readers should regard HumanNorm as a concurrent work, due to the released date. The reason we chose these two works for comparison is because they are the state-of-the-art approaches for text-to-3D realistic human generation. The other works, such as DreamWaltz~\cite{dreamwaltz} (NeurIPS'23) and TADA~\cite{tada} (3DV'24), are generally limited to cartoon-styled avatar with artifacts.
\section{Implementation details: Clean-NeRF} 
We implement Clean-NeRF~\cite{cleannerf} with TensoRF~\cite{tensorf} as the backbone. We use the vector-matrix (VM) decomposition model, and the hyper-parameters are defaulted as the suggested values of custom dataset in the official implementation, \eg, $\text{TV weight density}=0.1$, $\text{TV weight app}=0.01$.
\section{Implementation details: text inputs}
This section aims to provide more implementation details about the text inputs of diffusion models. To enhance the generating quality, we add some custom prompts in this work as follows:

\vspace{.75mm}
\noindent{\bf Positive prompts.} \quad We append the prompt \textit{``, blank background, (high quality), (best quality), (8k), masterpiece, whole body''} after the text description of each image. Besides, depending on the camera poses, custom direction-corresponding prompts such as \textit{``(side view), side face''}, \textit{``(front view), clear face''} and \textit{``(back view)''} are placed at the end of each positive prompt.

\vspace{.75mm}
\noindent{\bf Negative prompts.}\quad We follow the suggested negative prompt in the \textit{Realistic\_Vision\_V3.0\_VAE} model, namely, \textit{``(deformed iris, deformed pupils, semi-realistic, cgi, 3d, render, sketch, cartoon, drawing, anime:1.4), text, close up, cropped, out of frame, worst quality, low quality, jpeg artifacts, ugly, duplicate, morbid, mutilated, extra fingers, mutated hands, poorly drawn hands, poorly drawn face, mutation, deformed, blurry, dehydrated, bad anatomy, bad proportions, extra limbs, cloned face, disfigured, gross proportions, malformed limbs, missing arms, missing legs, extra arms, extra legs, fused fingers, too many fingers, long neck''}.

\section{Implementation details: hyper-parameters}
\vspace{.75mm}
\noindent{\bf Loss weight.}\quad Recall that we use $\mathcal{L}_\text{recon}$ and $\mathcal{L}_\text{sample}$ in PAR, where $\mathcal{L}_\text{sample}$ is associated with a random number $t$. In our experiments, we generate a random variable $t_i$ for each of the sampled views $w_i$, and the total loss is computed with the following weight:
\begin{equation}
    \mathcal{L}_\text{total} = \mathcal{L}_\text{recon}+\frac{1}{d}\sum^d_{i=1}t_i\mathcal{L}_\text{sample}\left(w_i\right)
\end{equation}

\vspace{.75mm}
\noindent{\bf Classifier-free guidance scale (CFG scale)}\quad controls the similarity between the input prompts and generated image of diffusion models. In the preprocessing stage, i.e., $\bG_1$ and $\bG_2$, this parameter is set at $4.5$, while our empirical experience suggests that the final results are not sensitive to this parameter. In IPAR, $\bG_3$ has a decreasing CFG scale from $4.5$ to $3.5$. In PAR, this parameter is set $3.5$.
%
%
\bibliographystyle{splncs04}
\bibliography{main}
\end{document}